\documentclass{article}

\usepackage{PRIMEarxiv}

\usepackage[utf8]{inputenc} 
\usepackage[T1]{fontenc}    
\usepackage{hyperref}       
\usepackage{url}            
\usepackage{booktabs}       
\usepackage{amsfonts}       
\usepackage{nicefrac}       
\usepackage{microtype}      
\usepackage{lipsum}
\usepackage{fancyhdr}       
\usepackage{graphicx}       
\graphicspath{{media/}}     
\usepackage{graphicx}
\usepackage{amsmath}
\usepackage{amssymb}
\usepackage{booktabs}
\usepackage{url}
\usepackage[caption=false, font=normalsize, labelfont=sf, textfont=sf]{subfig}

\usepackage{amsmath,amsfonts,bm}

\def\eqref#1{equation~\ref{#1}}

\def\1{\bm{1}}

\DeclareMathAlphabet{\mathsfit}{\encodingdefault}{\sfdefault}{m}{sl}
\SetMathAlphabet{\mathsfit}{bold}{\encodingdefault}{\sfdefault}{bx}{n}

\DeclareMathOperator*{\argmin}{arg\,min}

\title{Revisiting the Fragility of Influence Functions
}

\author{
  Jacob R. Epifano, Ravi P. Ramachandran \\
  Department of Electrical and Computer Engineering \\
  Rowan Univsercity \\
  Glassboro NJ, USA\\
  \texttt{epifanoj0@students.rowan.edu, ravi@rowan.edu} \\
   \And
  Aaron J. Masino \\
  Department of Biostatistics, Epidemiology, Informatics \\
  University of Pennsylvania Perelman School of Medicine \\
  Philadelphia PA, USA\\
  \texttt{aaron.masino@pennmedicine.upenn.edu} \\
   \And
  Ghulam Rasool \\
  Department of Machine Learning \\
  Moffitt Cancer Center \\
  Tampa FL, USA\\
  \texttt{ghulam.rasool@moffitt.org} \\
}

\begin{document}
\maketitle

\begin{abstract}
In the last few years, many works have tried to explain the predictions of deep learning models. Few methods, however, have been proposed to verify the accuracy or faithfulness of these explanations. Recently, influence functions, which is a method that approximates the effect that leave-one-out training has on the loss function, has been shown to be fragile. The proposed reason for their fragility remains unclear. Although previous work suggests the use of regularization to increase robustness, this does not hold in all cases. In this work, we seek to investigate the experiments performed in the prior work in an effort to understand the underlying mechanisms of influence function fragility. First, we verify influence functions using procedures from the literature under conditions where the convexity assumptions of influence functions are met. Then, we relax these assumptions and study the effects of non-convexity by using deeper models and more complex datasets. Here, we analyze the key metrics and procedures that are used to validate influence functions. Our results indicate that the validation procedures may cause the observed fragility. 
\end{abstract}


\section{Introduction}
\label{sec:intro}
Due to the black-box nature of Deep Neural Networks (DNNs), explaining the predictions of these models remains a challenging problem. Several techniques for addressing this challenge have been proposed such as saliency maps\cite{simonyan2014deep}, influence functions \cite{koh2017understanding}, concept activation vectors \cite{kim2018interpretability}, and activation atlases \cite{carter2019activation}. These techniques are not without problems. The fragility of these methods have been well studied, but few works have tried to understand where these methods break down \cite{ghorbani2019interpretation, basu2020influence}.\footnote{Code and Raw output files are available at the following url: \url{https://github.com/jrepifano/xai_is_fragile}}

Influence functions were originally proposed to diagnose and debug linear models by predicting the parameter or loss change due to removing a training instance \cite{cook1982residuals}. Their extension to deep learning models, however, did not occur until recently \cite{koh2017understanding}. Influence functions and their applications have been well studied since their reemergence and have since been adopted as a mainstream tool for the interpretation of deep models in a variety of data modalities \cite{cohen2020detecting, guo2020fastif,lee2020learning, han2020explaining}, including high-risk areas such as mortality prediction for patients in the Intensive Care Unit \cite{epifano2020towards}. Due to the diversity of the use cases for influence functions, understanding their limitations is imperative if they are to be used to explain model behavior. Without key validation procedures, we run the risk of providing misleading or incorrect information to the model users. 

To validate these methods, we must first agree on a metric to rate explanations. Spearman correlation between the approximate and true loss differences has been used as a metric to determine the accuracy of influence estimates. The approximate loss differences are given by the influence functions and the true loss differences are obtained by retraining an already trained network after removing a specific training sample \cite{koh2017understanding}. Recent works have used this metric to study the effects that increases in model and dataset size have on the influence functions. It has been found that influence functions are extremely sensitive to these increases \cite{basu2020influence}. 

It is well known that increases in model and dataset size affect the curvature of the loss function \cite{sagun2016eigenvalues, sagun2017empirical, ghorbani2019investigation, alain2019negative}. Convexity of the loss function is a critical assumption of influence functions as they heavily rely on the approximation of the inverse Hessian-vector product. The stochastic estimation algorithm used to compute the inverse Hessian-vector product assumes that the Hessian is positive semidefinite \cite{pearlmutter1994fast, agarwal2017second}.  Preliminary work has been done to try to remedy these problems via higher-order approximations \cite{koh2019accuracy} and group influences \cite{basu2020second}, i.e., computing loss differences for more than one training instance at a time. 

\begin{figure*}[t]
    \centering
    \subfloat{
    \includegraphics[width=0.45\textwidth]{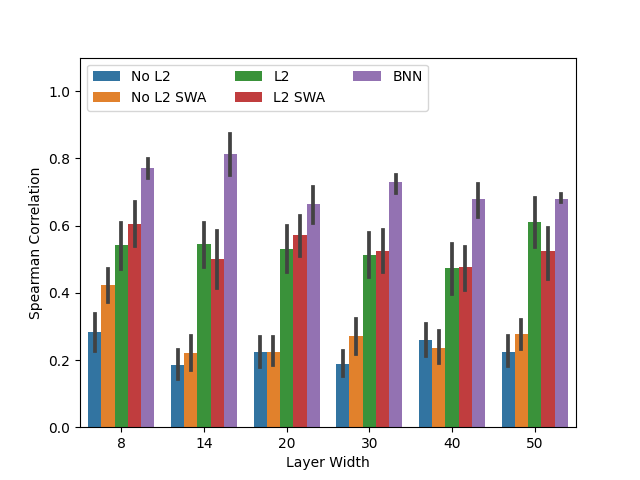}
    }
    \hfill
    \subfloat{
    \includegraphics[width=0.45\textwidth]{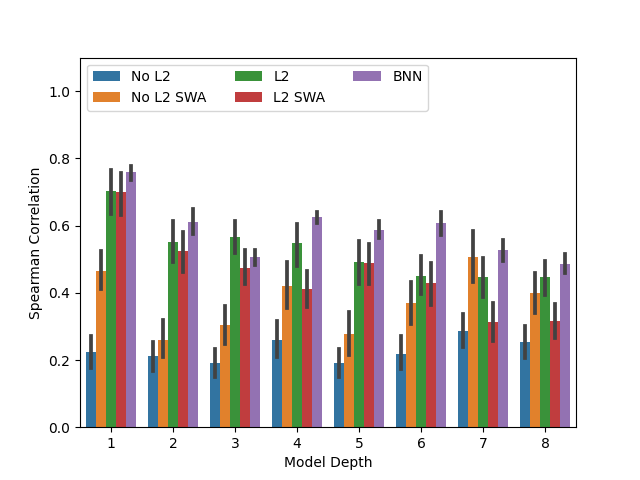}
    }
    \caption{Influence function performance evaluation on Iris dataset. \textbf{Left}: constant depth experiment. \textbf{Right}: constant width experiment. Spearman correlation between the true and approximate loss differences is on the y-axis (higher is better). The error bars represent the 95\% intervals obtained by repeating the experiment 50 times. \textbf{Blue}: training without weight decay. \textbf{Orange}: training without weight decay but with Stochastic Weight Averaging (SWA). \textbf{Green}: training with weight decay. \textbf{Red}: training with weight decay and SWA. \textbf{Purple}: training with BNN. We observe that the influence functions that come from BNN are significantly better than the rest of the methods in almost all cases. SWA has significant performance increases without the presence of regularization but with regularization has little effect and was removed for clarity. Statistical testing using one-way ANOVA revealed no significant difference ($p>.05$) between correlation values for any of the model types.}
    \label{fig:iris}
\end{figure*}

When discussing fragility, we must look at the whole system, not just the method in question. Deep neural networks have been shown to be sensitive to small perturbations via the weight initialization or by the order in which the data is given to the model \cite{smilkov2017smoothgrad, madhyastha2019model}. The problem lies in the noisy nature of the gradients. This problem has been linked to poor model convergence as well as explainability and attempts to address it include Gaussian averaging \cite{smilkov2017smoothgrad}, Stochastic Weight Averaging (SWA) \cite{izmailov2018averaging,madhyastha2019model} and model averaging through Bayesian Inference \cite{blundell2015weight}.

In this paper, we examine the cases where influence functions seemingly fail, i.e. have low Spearman correlation between approximate and true loss differences. We obtain the operands for the correlation using the retraining procedure introduced in \cite{koh2017understanding}, where the approximate loss differences are computed for the test point with the maximal loss using influence functions. Each training point is removed one at a time and the neural network is retrained from the optimal parameters until convergence in order to obtain the true difference in the loss function values. We determined that this training procedure is not valid for most applications of deep learning and present evidence for these cases.

\section{Background}
\subsection{Influence Functions}
Consider a standard classification problem where a label $y$ is predicted for each feature vector $x$. Let $z_i = (x_i, y_i)$, where $i = 1, 2, ..., N$, for $N$ instances in the dataset. It is assumed that we have a trained model where $\theta$ represents the trained network parameters. Our loss function can be written as $L(z, \theta) = \sum_{i=1}^N L(z_i, \theta)$. Our optimal model parameters are the set of parameters that minimize the loss: $\hat{\theta} = \argmin_{\theta \in \Theta}\sum_{i=1}^N L(z_i, \theta)$
\cite{koh2017understanding}. 
Koh \emph{et al.} \cite{koh2017understanding} offers insight on how to approximate the effect that removing a training point $z$ has on the parameters $\theta$. We compute the parameter change with $z$ upweighted by a small value, $\epsilon$. Using this upweighting scheme we obtain a new set of parameters, $\hat{\theta}_{\epsilon, z} = \textrm{arg}\min_{\theta \in \Theta}\frac{1}{n}\sum_{i=1}^n L(z_i, \theta)+\epsilon L(z, \theta)$
\cite{koh2017understanding}. 
Cook \emph{et al.} \cite{cook1982residuals} has shown that as $\epsilon$ approaches zero the influence of $z$ on the parameters is:
\begin{equation}
    \mathcal{I}_{\textrm{up,params}}(z)=\frac{d\hat{\theta}_{\epsilon, z}}{d\epsilon}\bigg\rvert_{\epsilon=0}= -H^{-1}_{\hat{\theta}}\nabla_\theta L(z,\hat{\theta}),
    \label{up_params}
\end{equation}
where $H_{\hat{\theta}} = \frac{1}{n}\sum_{i=1}^n \nabla^2_\theta L(z_i, \hat{\theta})$ is the Hessian. If we let $\epsilon = -\frac{1}{n}$, then we can approximate the parameter change as $\hat{\theta}_{-z} - \hat{\theta} \approx -\frac{1}{n}\mathcal{I}_{\textrm{up,params}}(z)$ \cite{koh2017understanding}. To study the effect of removing a training point on a test point $z_{\textrm{test}}$ on the loss function, we apply the chain rule: \cite{koh2017understanding}:
\begin{equation}
    \mathcal{I}_{\textrm{up,loss}}(z, z_{\textrm{test}}) = -\nabla_\theta L(z_{\textrm{test}}, \hat{\theta})^TH_{\hat{\theta}}^{-1}\nabla_\theta L(z, \hat{\theta})
    \label{up_loss}
\end{equation}

\begin{figure}[t]
    \centering
    \includegraphics[width=.8\textwidth]{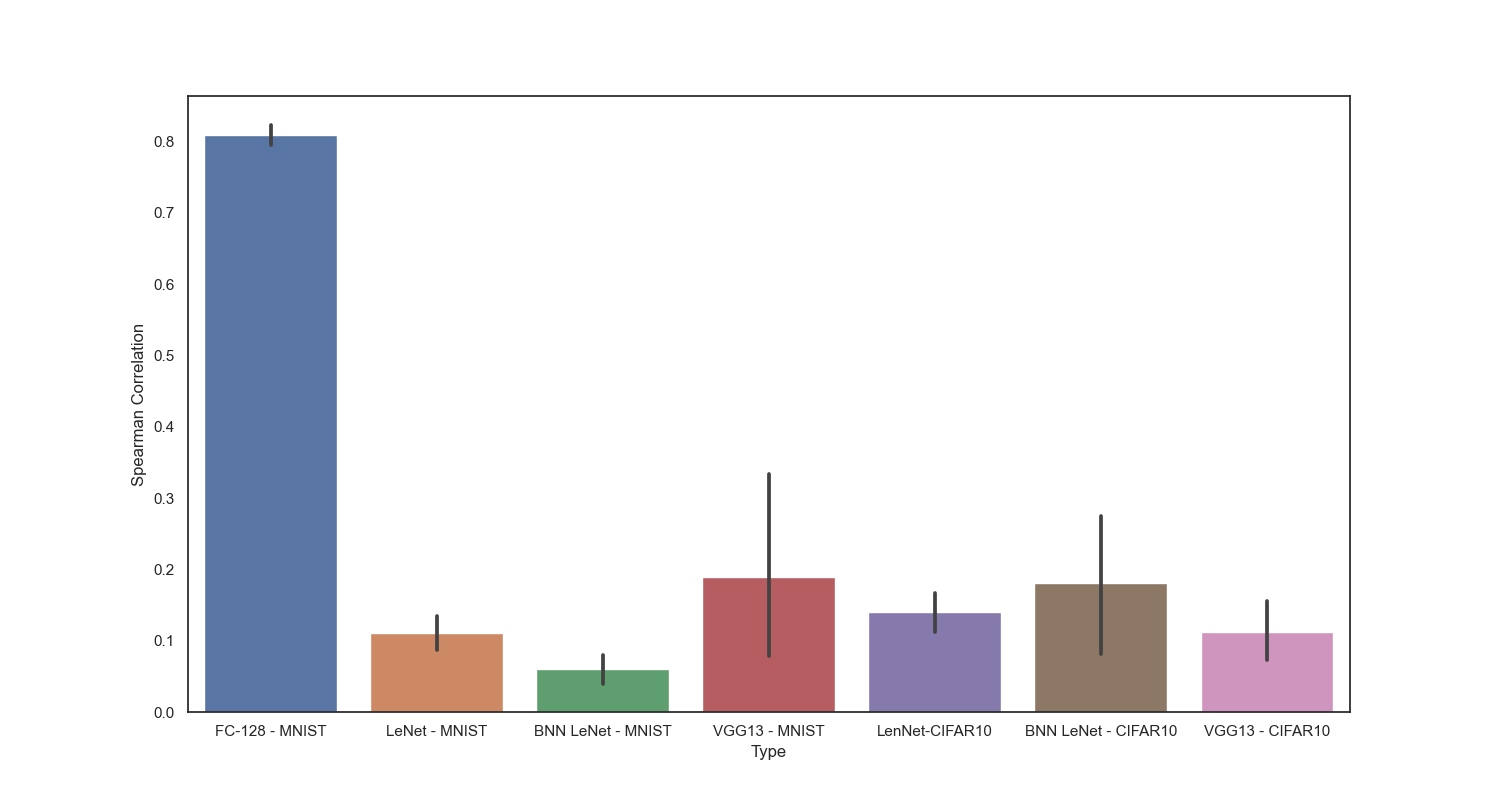}
    \caption{Spearman correlation between the true and approximate loss differences is on the y-axis (higher is better). The error bars represent the 95\% intervals obtained by repeating the experiment 10 times. We observe that the spearman correlation is only significant in the small fully connected model on the MNIST dataset. As the number of parameters increases, the influence function performance falls off sharply, which was expected. There are no significant differences between VDP and the other large models.}
    \label{fig:deep}
\end{figure}

\subsection{Influence Function Guidance}
Ideally, a model must be trained until the optimal parameters $\hat{\theta}$ are obtained in order to compute the influence functions. For a single test instance, $z_{\textrm{test}}$, we would then compute the inverse Hessian-vector product, $\nabla_\theta L(z_{\textrm{test}}, \hat{\theta})^TH_{\hat{\theta}}^{-1}$, using stochastic estimation \cite{pearlmutter1994fast}. In reality, due to non-linearities in our networks, our objective function may become non-convex and we obtain our parameters $\Tilde{\theta}$ via SGD, where $\Tilde{\theta} \neq \hat{\theta}$. In this case, the Hessian may have negative eigenvalues which would cause the stochastic estimation algorithm to not converge. To address this, we adopt a regularization scheme similar to L2 regularization discussed by Koh \emph{et al.}\cite{koh2017understanding}. We regularize the computation of the Hessian-vector product using a damping term of $\lambda = 0.01$. We can then compute the gradient of the loss as $\nabla_\theta L(z, \hat{\theta})$. The inner product of the Hessian-vector product and the gradient of the training instance results in a scalar value that tells us the approximate change in loss to expect on $z_{\textrm{test}}$ if we were to remove the training instance $z$. Note that we compute the gradient of the loss function with respect to \textbf{only} the parameters of the last layer \cite{koh2017understanding}.

\subsection{Non-convexity and Eigenvalues of the Hessian}
Due to the importance of the Hessian in the computation of influence functions, the convexity of the loss function and its effects on the Hessian are important. Recall that influence functions assume the Hessian is positive definite such that it is invertible. Koh \emph{et al.} \cite{koh2017understanding} have shown that even with negative Hessian eigenvalues it is still possible to obtain good influence estimates . It is understood that large overparameterized networks affect the convexity of the loss function \cite{sagun2017empirical, ghorbani2019investigation}, which we observe via the eigenvalues of the Hessian. Basu \emph{et al.} \cite{basu2020influence} have shown that larger eigenvalues are correlated with decreases in the Spearman correlation metric when network depth and width are increased. This contradicts the literature where the long tail of the Hessian Eigen Spectral Density (ESD) has been well studied for large DNNs and it has been shown that the largest eigenvalue does not tend to increase as width of the network increases \cite{sagun2016eigenvalues}. In this paper, we utilize a method developed by Yao \emph{et al.} \cite{yao2020pyhessian} to compute the eigenvalues of the Hessian in an effort to quantify the effect if any, of non-convexity and non-convergence on Influence functions.


\subsection{Bayesian Deep Neural Networks}
The current state of the art for influence functions, suggests that by applying L2 regularization to our networks during training, we can reduce the negative effects that are associated with overparameterization \cite{basu2020influence}. Variational Bayesian Learning has been shown to result in superior regularization, better model averaging and built-in uncertainty prediction \cite{blundell2015weight}. We select this method specifically for its regularization strength.


In this subsection, we present a modified version of the Extended Variational Inference model proposed by Dera \emph{et al.} \cite{dera2019extended}. We assume the covariance is zero and only propagate variance for each parameter.

For a given classification problem, we want to estimate the posterior distribution of the weights given the data, i.e., $p(\theta | D)$. This, however, is intractable due to the high dimensionality of the parameter space. We can approximate the true posterior by defining a variational distribution $q(\theta)$, which is assumed to be Gaussian. Since we want the variational distribution to be close to the true posterior, we minimize the Kullback - Leibler (KL) divergence. 
\begin{equation}
\begin{split}
    \argmin_\theta \textrm{KL}(q(\theta)||p(\theta|D)) = &-E_{q(\theta)}\left[\log p(D|\theta) \right] \\
    &+ \textrm{KL}(q(\theta)||p(\theta))
\end{split}
\end{equation}

To quantify the loss for the variational learning approach, we use the Evidence Lower Bound (ELBO), $\mathcal{L}(\theta, D)$ which consists of two parts, namely, the expected log-likelihood of the training data given the weights and a regularization term,
\begin{equation}
    \mathcal{L}(\theta, D) = E_{q(\theta)}[\log p(D|\theta)] - \textrm{KL}[q(\theta)|p(\theta)]
\end{equation}
where $\theta$ represents the weights of the network and $D$ represents the data label pairs.
Continuing the derivation gives
\begin{equation}
    \begin{split}
    \mathcal{L}(\theta, D) &= \frac{1}{N}\sum_{i=1}^N \log \left(\prod \sigma^2_{\hat{y}}\right)\\
    &+ \frac{1}{2}\sum_{i=1}^N ((\hat{y} - \mu_{\hat{y}})^2(\sigma^2_{\hat{y}})^{-1})\\
    &+ \frac{1}{2} \sum_{n=1}^l (j \log \sigma^2_l - ||\mu_l||^2_F - j\sigma^2_l)
    \end{split}
\end{equation}
where: $\hat{y}$ is the label, $\mu_{\hat{y}}$ is the output mean, $\sigma^2_{\hat{y}}$ is the output variance,  $l$ is the number of hidden layers and $j$ is the number of nodes in that layer.

\begin{figure*}[t]
    \centering
    \subfloat{
    \includegraphics[width=.45\textwidth]{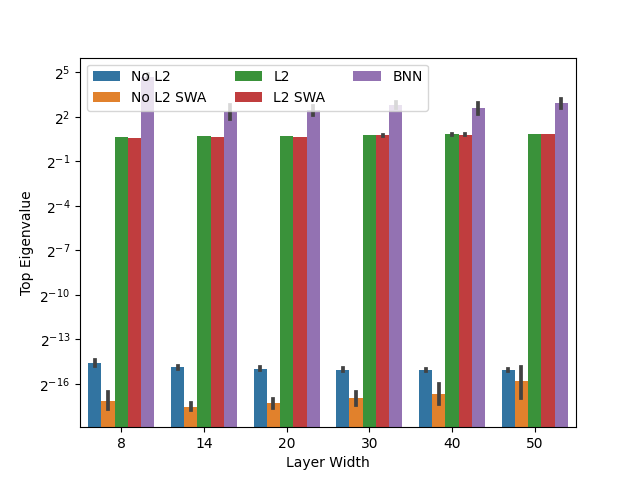}
    }
    \hfill
    \subfloat{
    \includegraphics[width=.45\textwidth]{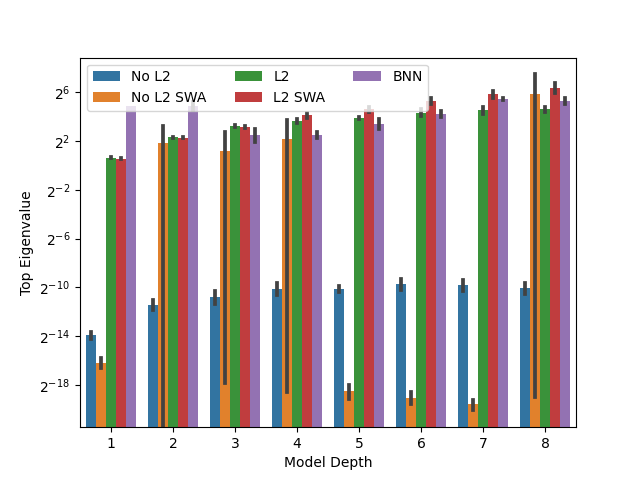}
    }
    \caption{Influence Function performance evaluation on Iris dataset. \textbf{Left}: constant depth experiment. \textbf{Right}: constant width experiment. Largest Eigenvalue is on the y-axis. The error bars represent the 95\% intervals obtained by repeating the experiment 50 times. \textbf{Blue}: training without weight decay. \textbf{Orange}: training without weight decay but with Stochastic Weight Averaging (SWA). \textbf{Green}: training with weight decay. \textbf{Red}: training with weight decay and SWA. \textbf{Purple}: training with BNN. This figure shows little correlation between curvature of the loss function. Statistical testing using one-way ANOVA showed no difference in the top eigenvalue for any model in the width or depth experiments.}
    \label{fig:iriseig}
\end{figure*}

For a neural network, we define the equations to propagate the first two moments. First, we assume the input $x$ is deterministic. The weights and biases of the first layer, $w$ are assumed to be Gaussian. If the incoming input is deterministic (input layer), then the output, $z$ is:
\begin{align}
    \mu_{z} &= \mu_{w}^Tx + \mu_{w_b} \label{muz}\\
    \sigma^2_{z} &= x^2[\sigma^2_{w}]^T + \sigma^2_{w_b} \label{sigz}
\end{align}
To propagate the moments through an arbitrary element-wise non-linear function, $f$ (e.g., ReLU, SELU), we use a first order Taylor-series approximation, to get an output, $a$:
\begin{align}
    a & = f(z) \label{eq:ZZXXXX} \\
    \mu_a &\approx f(\mu_z) \label{eq:aaaaaa} \\
    \sigma^2_a & = \sigma^2_z \odot (f'(\mu_z))^2
\end{align}

If the incoming input, $a$ is a random variable (for the intermediate layers), then the first two moments, $\tilde{y}$ are:
\begin{align}
    \mu_{\tilde{y}} & = \mu_{v}^T\mu_{a} + \mu_{v_b} \label{eq:mean_y} \\
    \sigma^2_{\hat{y}} & = \sigma^2_{v}[\sigma^2_{a}]^T + \mu^2_{v}[\sigma^2_{a}]^T + \mu^2_{a}[\sigma^2_{v}]^T + \sigma^2_{v_b} \label{eq:sigma_y}  
\end{align}
To propagate through a non-linear function that is not element-wise, e.g. softmax, we use a first order Taylor-series approximation \cite{simon2006optimal}. The output, $\hat{y}$ of the non-linear function, $g$ is:
\begin{align}
    \mu_{\hat{y}} &\approx g(\mu_{\tilde{y}}) \label{muyhat}\\
    \begin{split}
        \sigma^2_{\hat{y}} &\approx \mathbf{J}^2_g \odot \sigma^2_{\tilde{y}} \label{sigyhat}\\
        &\approx (\mu_{\hat{y}}(1-\mu_{\hat{y}}))^2 \odot \sigma^2_{\tilde{y}}
    \end{split}
\end{align}
where $\mathbf{J}_g$ is the Jacobian of $g$.

\section{Experiments}
\subsection{Iris Dataset}
To study the effect of random initialization on influence function estimates, we reproduced an experiment from Basu \emph{et al.} \cite{basu2020influence} using the Iris dataset. The Iris dataset consists of 150 instances with 4 features and 3 classes. The decision to use this dataset as a benchmark is due to its simplicity. To make our models more robust to random initialization, we considered weight decay as well as Stochastic Weight Averaging (SWA) and Bayesian Neural Networks (BNNs) as novel additions to this experiment \cite{madhyastha2019model, izmailov2018averaging, dera2019extended}.

This experiment was repeated for two types of DNNs: (1) DNNs with constant width (number of nodes in a hidden layer) and variable depth (number of hidden layers), and (2) DNNs with constant depth and variable width. In the experiments with variable depth, the number of nodes per hidden layer was held constant at 5 as in Basu \emph{et. al.} \cite{basu2020influence}. In the variable width experiments, the depth of the network was held constant at 1, i.e., one hidden layer only. We used the Adam optimizer with an initial learning rate of 0.001 as in Basu \emph{et. al.} \cite{basu2020influence}. A learning rate scheduler was used to decrease the learning rate by a factor of 10 if the loss did not decrease for 100 epochs. For the experiments with weight decay, we used a constant value of 0.005 as in Basu \emph{et. al.} \cite{basu2020influence}. Each experiment was repeated 50 times.

Koh \emph{et al.}\cite{koh2017understanding} showed that fine-tuning a trained DNN from the optimal parameters is approximately equal to retraining the same network with a training instance removed. Therefore, to obtain the true differences in loss when removing a test point, we replicate the training procedure outlined by Basu \emph{et al.} \cite{basu2020influence}. The models are initially trained for 60k epochs of full-batch gradient descent instead of SGD. The training instances are then sorted by their loss and the 40 training instances with the maximal loss are identified. We then fine-tune only the top layer for 7.5k epochs when individually removing each of the training points with the highest loss. Finally, we compute the influence function estimates for those training instances with respect to the test instance with the highest loss. The Spearman correlation between the true and approximate differences in loss are then computed. The eigenvalues of the Hessian for each network were computed via power iteration using the PyHessian Python package \cite{yao2020pyhessian}.

\begin{figure*}[t]
    \centering
    \subfloat{
    \includegraphics[width=.45\textwidth]{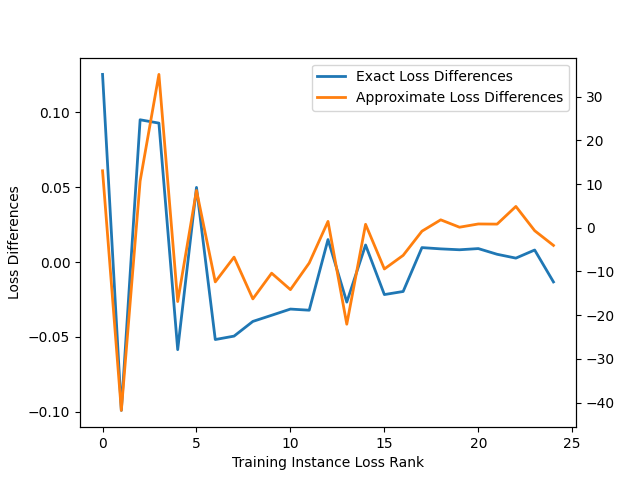}
    }
    \hfill
    \subfloat{
    \includegraphics[width=.45\textwidth]{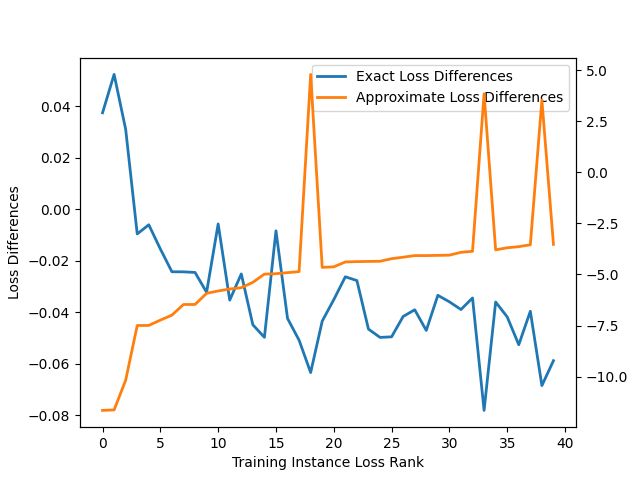}
    }
    \caption{Example of miss-relation. \textbf{Left:} Depth 1 width 5 network with weight decay on Iris dataset. \textbf{Right:} Small FC network with 128 nodes on MNIST dataset. We show that when the loss function is convex, our estimates match the true loss differences (Left, scale does not matter). When the loss function is non-convex there is significant deviation from the true loss differences. Both left and right receive an absolute spearman correlation of 0.85, which results in a miss-relation for the right graph.}
    \label{fig:lossdiff}
\end{figure*}

\subsection{MNIST and CIFAR10}
We drastically increase the model and dataset size to study the performance of influence functions in non-convex settings. Similar to the experiment described in Basu \emph{et al.} \cite{basu2020influence}, we chose to look at a small fully connected network, LeNet, and VGG13. Each model was trained in a similar manner as our previous experiment. The Adam optimizer was used with an initial learning rate of 0.001 and weight decay of 0.001. The learning rate was reduced by a factor of 10 if the loss did not decrease after 2 epochs. The test instance with the maximal loss was used to compute the influence functions and influence functions were computed for all training instances. We deviate from our previous experiment when choosing the training instances to remove and retrain. The true loss difference was computed for the top 40 most influential training points (highest absolute value) using the re-train from optimal approximation. The Spearman correlation between the true and estimated differences in loss was computed.

\subsection{Statistical Analysis}
We use one-way analysis of variance (ANOVA) to compare various dependent variables and establish statistical significance in various experiments described above.

\section{Results and Discussion}
\subsection{Effect of Model Size on the Influence Function Estimates}
In Figure \ref{fig:iris}, we present the Spearman's rank correlation coefficient ($\rho$) between the true and estimated loss differences for the Iris dataset for a variety of model types and sizes. We present four different types of models, including a model with L2 regularization, a model without L2 regularization, a model with SWA, and a BNN. The figure presents models trained using an increasing number of neurons in one layer (Figure \ref{fig:iris}-A) and increasing number of layers with fixed number of neurons in each layer (Figure \ref{fig:iris}-B). The true loss difference is found using the re-training strategy and the estimated loss difference is found using equation \ref{up_loss}. The error bars represent the 95\% confidence intervals obtained by repeating the experiment 50 times. It is evident from both sub-figures that for any type of model (L2, No-L2, SWA, and BNN), there is a minimal effect of increasing number of neurons or number of layers on the quality of estimate (of the influence of a training point on the selected test data point) provided by influence functions (using equation \ref{up_loss}). A statistical analysis performed using ANOVA did not reveal any significant effect of number of neurons or layers on the Spearman correlation ($p>0.5$ for all cases). Previously, Basu \emph{et al.} \cite{basu2020influence} had reported increasing model size (depth and width) degrades influence function estimates. We believe that the discrepancy between the reported results is linked to statistical rigor as no statistical tests or analyses were reported by Basu \emph{et al.} \cite{basu2020influence} to establish the effect of model size on the quality of estimates produced by influence functions.

We also observe that the estimates provided by influence function are more accurate for models with regularization, as shown in Figure \ref{fig:iris}, in particular, the Bayesian models (BNNs) outperform all other methods. We consider that the observed behavior is linked to (1) the ``ensemble" or ``average" effect introduced by Bayesian approaches in the model training, and (2) the type of regularization present in the ELBO loss function which has been shown to give these models superior self-compression properties \cite{carannante2020self}. This performance increase, however does not seem to carry over to our experiments with larger datasets (Figure \ref{fig:deep}), where all models were trained with regularization. This is congruent with the results obtained by Basu \emph{et al.} \cite{basu2020influence} on the same datasets.


\subsubsection{The Largest Eigen Value}
In Figure \ref{fig:iriseig} we observe the same trend that Basu \emph{et. al.} \cite{basu2020influence} found in the Iris experiment, e.g., the eigenvalues of the Hessian increase with model width and depth (ANOVA $p<0.05$). We do not however, relate the supposed decrease in influence function estimates to the increasing top eigenvalue as a proxy for curvature of the loss function given that our statistical results from Figure \ref{fig:iris} show that there are no significant differences between model sizes and influence function performance. Given that Koh \emph{et al.} \cite{koh2017understanding} have shown that even when most assumptions about convexity of the loss function have been broken, i.e., the optimal parameters have not been obtained ($\tilde{\theta} \neq \hat{\theta}$) and the Hessian has negative eigenvalues (Hessian not PD), we can still obtain ``good" influence estimates. We postulate that the problem lies with the methods that have been used to evaluate influence functions.



\begin{figure*}[t]
    \centering
    \subfloat{
    \includegraphics[width=.45\textwidth]{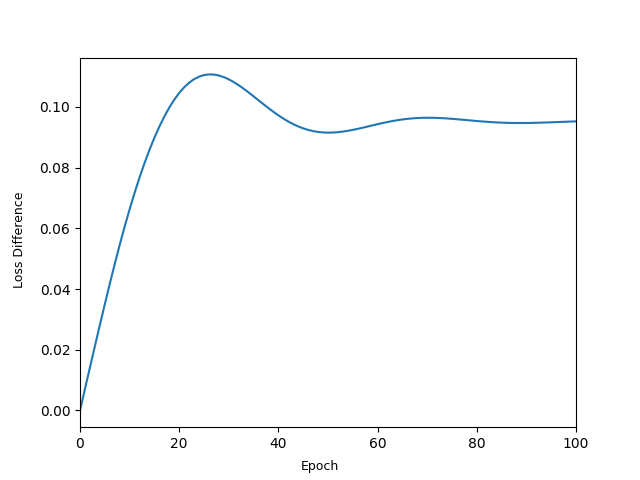}
    }
    \hfill
    \subfloat{
    \includegraphics[width=.45\textwidth]{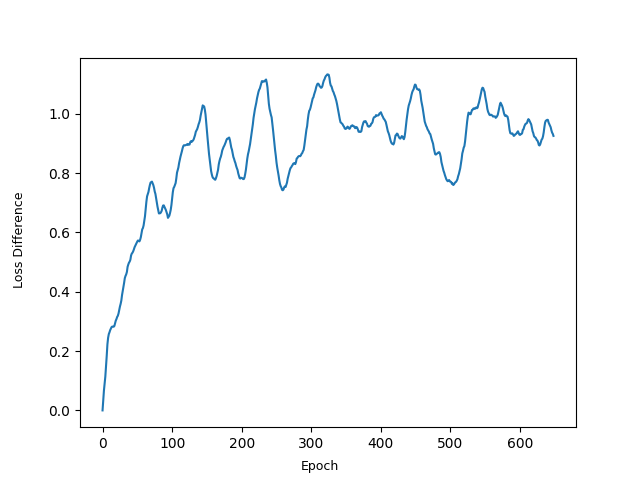}
    }
    \caption{The loss trajectories followed during re-raining loss. \textbf{Left} Depth 1 width 5 network with weight decay on Iris dataset. \textbf{Right} Small FC network with 128 nodes on MNIST dataset. Here we show the test loss as a function of re-training in convex and non-convex settings. The sharp jumps of the right plot indicate that the model leaves the minima that it settled in previously, which breaks the assumptions of the influence functions.}
    \label{fig:landscape}
\end{figure*}

\subsection{The (In-)validity of Spearman's Rank Correlation Coefficient} Spearman's rank correlation coefficient is an established metric for determining the accuracy of influence function estimates \cite{koh2017understanding, basu2020influence, basu2020second}. We note that the output of Equation \ref{up_loss} is the difference in the loss function value for the test instance if the training instance is removed. This loss difference can be positive or negative. For a training instance to be influential, it needs to have a large magnitude. 

In Figure \ref{fig:lossdiff}, we provide an example where Spearman's correlation coefficient is unable to capture the underlying relationship between the true loss difference and estimated loss difference, where the estimate is being calculated using equation \ref{up_loss}. The horizontal axis in both sub-figures (Figure \ref{fig:lossdiff} \textbf{left} and \textbf{right}) corresponds to the rank of the training point, where the rank is determined by the approximate loss difference. Thus, we should expect to see the exact loss differences (blue points) move from a large magnitude towards zero as we move from left to right on the horizontal axis. In Figure \ref{fig:lossdiff} (a), the estimated and true values (after ignoring the scale) are close to each other. In Figure \ref{fig:lossdiff} (b), the values of true and estimated loss differences are significantly different from each other. However, the value of Spearman's correlation coefficient for both cases is approximately 0.85. 

We consider that since the relationship between the estimated and true loss function difference values may not always be \textbf{monotonically} decreasing or increasing, the Spearman's correlation can lead to misleading results. 



\subsection{Re-training for Optimal Parameters}
To compute the Spearman's correlation coefficient, we need to know the true difference in the loss function value. This requires retraining models for every training instance that we want to analyze. This is a very costly operation in time. The re-training from optimal parameters has been shown to be an approximately equivalent alternative to retraining from scratch \cite{koh2017understanding}. Previous works have not proven that this approximation is valid for large datasets \cite{koh2017understanding, basu2020influence}.  It has been well established that increasing model and dataset complexity increases the largest eigenvalue of the Hessian of the loss function \cite{sagun2017empirical, ghorbani2019investigation}. While we have demonstrated that the increasing curvature does not affect estimates made by influence functions with small datasets and models, with large datasets and models the extreme curvature of the loss function makes us question the validity of the re-training approximation \cite{zhang2018three}. To study this, we looked at the loss of the test instance at each epoch during re-training in both the Iris and MNIST experiments. In Figure \ref{fig:landscape}, the test loss difference is plotted against epochs on the horizontal axis. We note significant differences differences in the trajectories followed by the gradient descent algorithm for two cases (Iris - Figure \ref{fig:landscape} left and MNIST \ref{fig:landscape} right). The Iris model has a well damped convergence whereas the the MNIST model is underdamped and does not seem to converge as smoothly as did Iris. 

\subsection{The Effect of Large Networks}
We consider large neural networks as having more parameters, more non-linear operations owing to their depth, and reluctantly requiring large datasets for training. We note that originally Cook and Weisberg  derived influence functions for regression models, which can be considered as neural networks with one layer and mean-square error loss function \cite{cook1982residuals}. Recently, Koh \emph{et al.} \cite{koh2017understanding} extended the idea of using influence functions in deep neural networks by treating all but the last layer of the deep neural network as a feature extractor. The influence functions were computed with respect to only the last layer. This practice seems to work in some cases and produce promising results \cite{koh2017understanding}. However, it does not account for the large dimensionality of the final layer of modern neural networks. This proves to be a problem when these large parameter matrices become ill-conditioned \cite{belsley2005regression}. This problem was captured by Basu \emph{et al.} \cite{basu2020influence} in their analysis of large datasets like CIFAR-100 and ImageNet where true differences in losses from removing training instances resulted in very noisy results. 

Large neural networks may have more layers (depth) and/or more operations per layer (width). This results in increasing the number of non-linear operations which are performed on the data for calculating the loss function. The most popular implementation of influence functions, as defined by Koh \emph{et al.} \cite{koh2017understanding}, relies on only a first-order Taylor series approximation to efficiently compute influence (Eq. \ref{eq:taylor}). We argue that the increasing number of non-linear operations strongly affects the convexity assumption of loss function $R(\theta)$ (Eq. \ref{eq:risk}) as used in the mathematical relationships derived for influence functions (Eq. \ref{eq:h}). There is evidence suggesting that adding the second term of the Taylor series in the influence function approximation improves the estimates \cite{koh2019accuracy, basu2020second}.

Finally, large networks typically go hand-in-hand with large datasets. From equation \ref{up_params}, it is evident that removing a training instance is equivalent to up-weighting it by $\epsilon = - \frac{1}{n}$. In the Iris dataset, $|\epsilon| \approx 6.6e-3$ compared to MNIST where $|\epsilon| \approx 1.6e-5$. Any larger datasets will lead to smaller epsilon, that is:
\begin{align}
    \hat{\boldsymbol\theta}_{-z} - \hat{\boldsymbol\theta} \approx 0 \text{ or } \hat{\boldsymbol\theta}_{-z} \approx \hat{\boldsymbol\theta}.
\end{align}
In other words, owing to the large dataset, the influence of a single training point on a test sample is asymptotically approaching zero. Perhaps the first-order Taylor series approximation of the influence functions does not provide enough resolution to predict loss differences when predicting on only one training instance. If one wanted to use influence functions in large datasets like CIFAR-100 and ImageNet, one would have to turn to higher-order approximations of influence functions as well as group influences. In Basu \emph{et al. }\cite{basu2020second}, promising results were obtained by examining the effect of a second-order approximation as well as group influence. These solutions of course have an associated cost. Adding a second-order term increases the cost and complexity of the analysis. Optimal group selection is also a non-trivial and expensive operation.

While there is theoretical evidence to suggest that the first-order implementation of influence functions is fragile, due to the difficulty of finding robust ways to empirically evaluate them in difficult settings, their supposed fragility remains unclear.

\section{Limitations}
While we have established that the procedures used to measure the accuracy of influence functions are flawed in multiple ways, we have not been able to ascertain exactly where or why these procedures break down. It appears that the answer lies with increasing model and dataset size. To precisely define the boundaries on where violating the approximations that Koh \emph{et al.} \cite{koh2017understanding} have established is valid, we would need to exhaustively search the space of increasing complexity of the model and dataset.


\section{Conclusion}
Validating the performance of explanation methods is a key area of deep learning that has not been well studied. In particular, the validation of influence functions in deep learning has been an area of interest. In this work, we analyzed several experiments from the recent literature in order to understand the fragility of influence functions. We obtained results that conflict with those of our peers, which we attribute to the repetition in our experimental design as well as the misuse of the Spearman correlation metric and retraining procedure.

While we have demonstrated that the methods we use to measure the accuracy of influence functions are flawed, we must not conclude that influence functions are uninformative. Due to the flaw in validation methodology, we do not have any evidence to support the claim that the explanations provided by influence functions are not accurate or faithful to the original model. Future efforts should be focused on developing robust validation frameworks for explainable methods in order to foster user-model trust. 


\appendix
\section{Influence Function Derivation}
This derivation was taken directly from Koh \emph{et al.} \cite{koh2017understanding} and has been reproduced below:

Influence functions are considered one of the classic technique from robust statistics that can quantify the change in model parameters attributed to up-weighting a training point by an infinitesimal amount. In the following, we derive mathematical relationships for influence functions, examine their underlying assumptions, and attempt to explain these in the context of large neural networks. We start by defining an optimization problem where $\hat{\theta}$ minimizes the empirical risk following Koh \emph{et al.} \cite{koh2017understanding}:

\begin{equation}
R(\theta) \stackrel{\text{def}}{=} \frac{1}{n}\sum_{i=1}^n L(z_i, \theta).
\label{eq:risk}
\end{equation}

A fundamental assumption for influence functions is that $R$ is twice-differentiable and \textbf{strongly} convex in $\theta$. That is, the Hessian, as defined by:

\begin{equation}
H_{\hat{\theta}} \stackrel{\text{def}}{=} \nabla^2 R(\hat{\theta}) = \frac{1}{n}\nabla^2_\theta L(z_i, \hat{\theta}),
\label{eq:h}
\end{equation}

exists and is positive definite. The convexity assumption guarantees the existence of $H_{\hat{\theta}}^{-1}$. Recall, that we approximate the removal of a training point by up-weighting the parameters by a small quantity $\epsilon \approx -\frac{1}{n}$, where $n$ is the total number of training data points. The perturbed parameters, $\hat{\theta}_{\epsilon, z}$ can be written as \cite{koh2017understanding}:

\begin{equation}
\hat{\theta}_{\epsilon, z} = \textrm{arg min}_{\theta \in \Theta}\left[R(\theta) + \epsilon L(z, \theta)\right].
\label{eq:theta-z}
\end{equation}

The total parameter change by up-weighting a training example can be defined as $\Delta_\epsilon = \hat{\theta}_{\epsilon, z} - \hat{\theta}$. Differentiating with respect to $\epsilon$ and noting that $\hat{\theta}$ doesn't depend on $\epsilon$, we can write:

\begin{equation}
\frac{d \hat{\theta}_{\epsilon, z}}{d \epsilon} = \frac{d \Delta_\epsilon}{d \epsilon}.
\label{eq:deriv}
\end{equation}

We note that for the optimal parameters $\hat{\theta}_{\epsilon, z}$, we can rewrite eq. \ref{eq:theta-z} as:

\begin{equation}
0 = \nabla R(\hat{\theta}_{\epsilon, z}) + \epsilon \nabla L(z, \hat{\theta}_{\epsilon, z}).
\label{eq:opt}
\end{equation}

Since $\hat{\theta}_{\epsilon, z} \rightarrow \hat{\theta}$ as $\epsilon \rightarrow 0$, the \textbf{first-order} Taylor expansion of the right-hand side produces:

\begin{equation}
        0 \approx \left[ \nabla R(\hat{\theta}) + \epsilon \nabla L(z, \hat{\theta}) \right] + \left[ \nabla^2 R(\hat{\theta}) + \epsilon \nabla^2 L(z, \hat{\theta}) \right]\Delta_\epsilon.
        \label{eq:taylor}
\end{equation}

Solving for $\Delta_\epsilon$,

\begin{equation}
    \Delta_\epsilon = -\left[ \nabla^2 R(\hat{\theta}) + \epsilon \nabla^2 L(z, \hat{\theta}) \right]^{-1}\left[ \nabla R(\hat{\theta}) + \epsilon \nabla L(z, \hat{\theta}) \right]
    \label{eq:deltep}
\end{equation}

Since $\hat{\theta}$ minimizes $R$, then $\nabla R(\hat{\theta}) = 0$. Neglecting higher order $\epsilon$ terms,

\begin{equation}
    \Delta_\epsilon \approx - \nabla^2 R(\hat{\theta})^{-1}\nabla L(z, \hat{\theta})\epsilon.
    \label{eq:simplify}
\end{equation}

When we substitute Equations \ref{eq:h} and \ref{eq:deriv}, we have:

\begin{equation}
    \left . \frac{d \hat{\theta}_{\epsilon, z}}{d \epsilon} \right|_{\epsilon = 0} = -H_{\hat{\theta}}^{-1} \nabla L(z, \hat{\theta}) \stackrel{\text{def}}{=} \mathcal{I}_{\text{up,params}}(z).
\end{equation}


\subsubsection*{Acknowledgments}
\noindent Jacob R. Epifano is supported by US Department of Education GAANN award P200A180055. Ghulam Rasool was partly supported by NSF OAC-2008690.

 \bibliographystyle{unsrt} 
 \bibliography{references}

\begin{thebibliography}{10}

\bibitem{simonyan2014deep}
Karen Simonyan, Andrea Vedaldi, and Andrew Zisserman.
\newblock Deep inside convolutional networks: Visualising image classification
  models and saliency maps.
\newblock In {\em In Workshop at International Conference on Learning
  Representations}. Citeseer, 2014.

\bibitem{koh2017understanding}
Pang~Wei Koh and Percy Liang.
\newblock Understanding black-box predictions via influence functions.
\newblock In {\em International Conference on Machine Learning}, pages
  1885--1894. PMLR, 2017.

\bibitem{kim2018interpretability}
Been Kim, Martin Wattenberg, Justin Gilmer, Carrie Cai, James Wexler, Fernanda
  Viegas, et~al.
\newblock Interpretability beyond feature attribution: Quantitative testing
  with concept activation vectors (tcav).
\newblock In {\em International conference on machine learning}, pages
  2668--2677. PMLR, 2018.

\bibitem{carter2019activation}
Shan Carter, Zan Armstrong, Ludwig Schubert, Ian Johnson, and Chris Olah.
\newblock Activation atlas.
\newblock {\em Distill}, 4(3):e15, 2019.

\bibitem{ghorbani2019interpretation}
Amirata Ghorbani, Abubakar Abid, and James Zou.
\newblock Interpretation of neural networks is fragile.
\newblock In {\em Proceedings of the AAAI Conference on Artificial
  Intelligence}, volume~33, pages 3681--3688, 2019.

\bibitem{basu2020influence}
Samyadeep Basu, Phil Pope, and Soheil Feizi.
\newblock Influence functions in deep learning are fragile.
\newblock In {\em International Conference on Learning Representations}, 2020.

\bibitem{cook1982residuals}
R~Dennis Cook and Sanford Weisberg.
\newblock {\em Residuals and influence in regression}.
\newblock New York: Chapman and Hall, 1982.

\bibitem{cohen2020detecting}
Gilad Cohen, Guillermo Sapiro, and Raja Giryes.
\newblock Detecting adversarial samples using influence functions and nearest
  neighbors.
\newblock In {\em Proceedings of the IEEE/CVF Conference on Computer Vision and
  Pattern Recognition}, pages 14453--14462, 2020.

\bibitem{guo2020fastif}
Han Guo, Nazneen~Fatema Rajani, Peter Hase, Mohit Bansal, and Caiming Xiong.
\newblock Fastif: Scalable influence functions for efficient model
  interpretation and debugging.
\newblock {\em arXiv preprint arXiv:2012.15781}, 2020.

\bibitem{lee2020learning}
Donghoon Lee, Hyunsin Park, Trung Pham, and Chang~D Yoo.
\newblock Learning augmentation network via influence functions.
\newblock In {\em Proceedings of the IEEE/CVF Conference on Computer Vision and
  Pattern Recognition}, pages 10961--10970, 2020.

\bibitem{han2020explaining}
Xiaochuang Han, Byron~C Wallace, and Yulia Tsvetkov.
\newblock Explaining black box predictions and unveiling data artifacts through
  influence functions.
\newblock {\em arXiv preprint arXiv:2005.06676}, 2020.

\bibitem{epifano2020towards}
Jacob~R Epifano, Ravi~P Ramachandran, Sharad Patel, and Ghulam Rasool.
\newblock Towards an explainable mortality prediction model.
\newblock In {\em 2020 IEEE 30th International Workshop on Machine Learning for
  Signal Processing (MLSP)}, pages 1--6. IEEE, 2020.

\bibitem{sagun2016eigenvalues}
Levent Sagun, Leon Bottou, and Yann LeCun.
\newblock Eigenvalues of the hessian in deep learning: Singularity and beyond.
\newblock {\em arXiv preprint arXiv:1611.07476}, 2016.

\bibitem{sagun2017empirical}
Levent Sagun, Utku Evci, V~Ugur Guney, Yann Dauphin, and Leon Bottou.
\newblock Empirical analysis of the hessian of over-parametrized neural
  networks.
\newblock {\em arXiv preprint arXiv:1706.04454}, 2017.

\bibitem{ghorbani2019investigation}
Behrooz Ghorbani, Shankar Krishnan, and Ying Xiao.
\newblock An investigation into neural net optimization via hessian eigenvalue
  density.
\newblock In {\em International Conference on Machine Learning}, pages
  2232--2241. PMLR, 2019.

\bibitem{alain2019negative}
Guillaume Alain, Nicolas~Le Roux, and Pierre-Antoine Manzagol.
\newblock Negative eigenvalues of the hessian in deep neural networks.
\newblock {\em arXiv preprint arXiv:1902.02366}, 2019.

\bibitem{pearlmutter1994fast}
Barak~A Pearlmutter.
\newblock Fast exact multiplication by the hessian.
\newblock {\em Neural computation}, 6(1):147--160, 1994.

\bibitem{agarwal2017second}
Naman Agarwal, Brian Bullins, and Elad Hazan.
\newblock Second-order stochastic optimization for machine learning in linear
  time.
\newblock {\em The Journal of Machine Learning Research}, 18(1):4148--4187,
  2017.

\bibitem{koh2019accuracy}
Pang~Wei Koh, Kai-Siang Ang, Hubert~HK Teo, and Percy Liang.
\newblock On the accuracy of influence functions for measuring group effects.
\newblock {\em arXiv preprint arXiv:1905.13289}, 2019.

\bibitem{basu2020second}
Samyadeep Basu, Xuchen You, and Soheil Feizi.
\newblock On second-order group influence functions for black-box predictions.
\newblock In {\em International Conference on Machine Learning}, pages
  715--724. PMLR, 2020.

\bibitem{smilkov2017smoothgrad}
Daniel Smilkov, Nikhil Thorat, Been Kim, Fernanda Vi{\'e}gas, and Martin
  Wattenberg.
\newblock Smoothgrad: removing noise by adding noise.
\newblock {\em arXiv preprint arXiv:1706.03825}, 2017.

\bibitem{madhyastha2019model}
Pranava Madhyastha and Rishabh Jain.
\newblock On model stability as a function of random seed.
\newblock {\em arXiv preprint arXiv:1909.10447}, 2019.

\bibitem{izmailov2018averaging}
Pavel Izmailov, Dmitrii Podoprikhin, Timur Garipov, Dmitry Vetrov, and
  Andrew~Gordon Wilson.
\newblock Averaging weights leads to wider optima and better generalization.
\newblock {\em arXiv preprint arXiv:1803.05407}, 2018.

\bibitem{blundell2015weight}
Charles Blundell, Julien Cornebise, Koray Kavukcuoglu, and Daan Wierstra.
\newblock Weight uncertainty in neural network.
\newblock In {\em International Conference on Machine Learning}, pages
  1613--1622. PMLR, 2015.

\bibitem{yao2020pyhessian}
Zhewei Yao, Amir Gholami, Kurt Keutzer, and Michael~W Mahoney.
\newblock Pyhessian: Neural networks through the lens of the hessian.
\newblock In {\em 2020 IEEE International Conference on Big Data (Big Data)},
  pages 581--590. IEEE, 2020.

\bibitem{dera2019extended}
Dimah Dera, Ghulam Rasool, and Nidhal Bouaynaya.
\newblock Extended variational inference for propagating uncertainty in
  convolutional neural networks.
\newblock In {\em 2019 IEEE 29th International Workshop on Machine Learning for
  Signal Processing (MLSP)}, pages 1--6. IEEE, 2019.

\bibitem{simon2006optimal}
Dan Simon.
\newblock {\em Optimal state estimation: Kalman, H infinity, and nonlinear
  approaches}.
\newblock John Wiley \& Sons, 2006.

\bibitem{carannante2020self}
Giuseppina Carannante, Dimah Dera, Ghulam Rasool, and Nidhal~C Bouaynaya.
\newblock Self-compression in bayesian neural networks.
\newblock In {\em 2020 IEEE 30th International Workshop on Machine Learning for
  Signal Processing (MLSP)}, pages 1--6. IEEE, 2020.

\bibitem{zhang2018three}
Guodong Zhang, Chaoqi Wang, Bowen Xu, and Roger Grosse.
\newblock Three mechanisms of weight decay regularization.
\newblock {\em arXiv preprint arXiv:1810.12281}, 2018.

\bibitem{belsley2005regression}
David~A Belsley, Edwin Kuh, and Roy~E Welsch.
\newblock {\em Regression diagnostics: Identifying influential data and sources
  of collinearity}.
\newblock John Wiley \& Sons, 2005.

\end{thebibliography}

\end{document}